\documentclass[runningheads]{llncs}

\usepackage{eccv}

\usepackage{eccvabbrv}

\usepackage{graphicx}
\usepackage{booktabs}
\usepackage{multirow}
\usepackage{caption}
\usepackage{algorithm2e}
\RestyleAlgo{ruled}
\usepackage{algpseudocode}
\usepackage{blindtext}
\usepackage{wrapfig}

\usepackage[accsupp]{axessibility}  
\usepackage{amssymb}

\usepackage{pifont}
\newcommand{\cmark}{\textcolor{green}{\ding{51}}}%
\newcommand{\xmark}{\textcolor{red}{\ding{55}}}%

\newcommand{\argmin}{\operatorname*{argmin}}

\usepackage{hyperref}

\usepackage{orcidlink}
\newcommand{\minisection}[1]{\vspace{0.025in} \noindent {\bf #1}\ \ }

\renewcommand{\cmark}{\ding{51}}%
\renewcommand{\xmark}{\ding{55}}%

\begin{document}


\title{Exemplar-free Continual Representation Learning via Learnable Drift Compensation}

\titlerunning{Learnable Drift Compensation}

\author{Alex Gomez-Villa\inst{1,2} \and
Dipam Goswami\inst{1,2} \and
Kai Wang\inst{1}\thanks{Corresponding author}\and
Andrew D. Bagdanov\inst{4} \and
Bartlomiej Twardowski\inst{1,2,3}\and
Joost van de Weijer\inst{1,2}}

\authorrunning{A.~Gomez-Villa et al.}

\institute{Computer Vision Center, Barcelona, Spain \and
Universitat Autonoma de Barcelona, Barcelona, Spain \and
IDEAS NCBR, Warsaw, Poland \and
MICC, University of Florence, Florence, Italy \\
\email{\{agomezvi,dgoswami,kwang,btwardowski,joost\}@cvc.uab.es, andrew.bagdanov@unifi.it}
}

\maketitle

\begin{abstract}
Exemplar-free class-incremental learning using a backbone trained from scratch and starting from a small first task presents a significant challenge for continual representation learning. Prototype-based approaches, when continually updated, face the critical issue of semantic drift due to which the old class prototypes drift to different positions in the new feature space.
Through an analysis of prototype-based continual learning, we show that forgetting is not due to diminished discriminative power of the feature extractor, and can potentially be corrected by drift compensation. 
To address this, we propose Learnable Drift Compensation (LDC), which can effectively mitigate drift in any moving backbone, whether supervised or unsupervised. LDC is fast and straightforward to integrate on top of existing continual learning approaches. Furthermore, we showcase how LDC can be applied in combination with self-supervised CL methods, resulting in the first exemplar-free semi-supervised continual learning approach. We achieve state-of-the-art performance in both supervised and semi-supervised settings across multiple datasets. Code is available at \url{https://github.com/alviur/ldc}. 
\end{abstract}

\section{Introduction}
\label{sec:intro}
The standard supervised learning paradigm assumes that data are available in a single training session. This belies, however, the reality of many real-world scenarios in which machine learning methods must deal with data that arrive in a continuous and non-stationary stream of tasks. These shifting distributions lead to catastrophic forgetting if approached naively (for instance, with continuing stochastic gradient descent)~\cite{mccloskey1989catastrophic}. Continual Learning (CL) considers the design of algorithms capable of learning from such streams of non-stationary data. 

Several techniques have been proposed for continually training models, such as regularization~\cite{li2017learning,kirkpatrick2017overcoming, zenke2017continual, aljundi2018memory}, exemplar replay~\cite{rebuffi2017icarl, castro2018end, wu2019large, hou2019learning, douillard2022dytox}, and growing architectures~\cite{rusu2016progressive}. One group of methods focuses on the more difficult exemplar-free continual learning scenario where the storing of exemplars is prohibited~\cite{zhu2021prototype,zhu2022self,goswami2024fecam,petit2023fetril,yu2020semantic}. This setting is relevant in applications where storing previous-task samples is not feasible due to memory limitations, legal constraints arising from new data privacy regulations (like GDPR), or when handling sensitive data such as medical imaging where patient identity must be protected.

One of the main limitations of exemplar-free CL methods is that, as the backbone is updated during the training of new tasks, the network experiences semantic feature drift~\cite{yu2020semantic}. This feature drift results in catastrophic forgetting and a significant drop in performance. As a consequence, most exemplar-free methods are evaluated in Warm Start settings (starting from a large first task or from a pretrained network) to start from a strong backbone and thus avoid significant semantic drift of backbone network~\cite{yu2020semantic,goswami2024fecam,zhu2021prototype,zhu2021il2a,zhu2022self,mcdonnell2024ranpac,hou2019learning,wu2019large,liu2020generative}. In fact, a recent paper~\cite{petit2023fetril} showed that a simple strategy combined with a frozen backbone after learning the first task can lead to very competitive results in this Warm Start settings. In this paper, we focus on the exemplar-free Cold Start setting~\cite{magistri2024elastic} in which we train a network from scratch, starting with a small task, and continually update the backbone.\footnote{This setting is predominant in exemplar-based continual learning but hardly explored for exemplar-free methods.}

Class Prototype (CP) accumulation is among the most successful approaches to exemplar-free continual learning~\cite{goswami2024fecam,janson2022simple,mcdonnell2024ranpac,panos2023session}. CP methods store class prototypes computed using training data and then use them to classify using a distance classifier in feature space. Class Prototype methods are attractive since they can be agnostically applied to any feature extractor, such as multilayer perceptrons, convolutional neural networks, or transformers. 
CP methods have been successfully applied in Warm Start settings~\cite{goswami2024fecam,mcdonnell2024ranpac,petit2023fetril} but are expected to struggle in Cold Start settings due to significant semantic drift.

In this article, we aim to apply class prototype methods in Cold Start settings. The main challenge in this setting is the semantic drift of the feature backbone, due to which class prototypes become progressively invalidated since the mean position of each class moves with each new task and cannot be corrected without access to old data in exemplar-free settings. However, in an initial analysis of feature drift (see Fig.~\ref{fig:drift}), we observe that an oracle that perfectly compensates for feature drift leads to a significant recovery of performance. This is an important observation since it shows that semantic drift has not diminished the discriminative power of the feature representations. Consequently, we show that if drift is correctly estimated, feature drift compensation can correct for most of the performance drop resulting from it. 
Existing drift estimation methods~\cite{yu2020semantic,toldo2022bring} assume that the drift can locally be approximated with a translation. Instead, we propose a simple Learnable Drift Compensation (LDC) approach to update old class prototypes which has not prior assumptions on the nature of the drift.
The learnable compensation maps the previous task features to the current feature space via a forward projector network. 
This compensation can be learned from only the stored old class prototypes and does not require class labels. 
Owing to the label-agnostic nature of LDC, we propose the first semi-supervised exemplar-free approach, which uses the labeled data only to compute prototypes and all available data to learn a robust forward projector.
The main contributions of this work are as follows:
\begin{itemize}
    \item We show that a large part of forgetting in prototype-based methods can be addressed by prototype compensation. Very little actual forgetting of features crucial for previous tasks is occurring, which motivates the need to develop drift compensation methods for continual representation learning.
    \item We propose a new exemplar-free continual learning method, which we call Learnable Drift Compensation (LDC), that is based on class prototype accumulation, can compensate the drift of a moving backbone, and is fast and easy to add on top of CP-based methods.
    \item We experimentally demonstrate that LDC is better than previous drift compensation methods for supervised exemplar-free CL setup. We show that LDC can be applied to any self-supervised CL method to create the first, to the best of our knowledge, exemplar-free semi-supervised CL method. 
    
\end{itemize}

\section{Related Work}

Continual Learning~\cite{mccloskey1989catastrophic,kirkpatrick2017overcoming} addresses the issue of \textit{catastrophic forgetting} when training model on a non-i.i.d stream of data~\cite{masana2022class}. Various methods~\cite{masana2022class, zhou2023pycil} have been proposed to tackle this problem, including parameter isolation methods~\cite{rusu2016progressive,serra2018overcoming, mallya2018packnet}, regularization techniques~\cite{aljundi2018memory,kirkpatrick2017overcoming,li2017learning}, and rehearsal strategies~\cite{rebuffi2017icarl,chaudhry2018efficient}. In this paper, we  focus on class-incremental learning~\cite{vandeven2019three}, where the learner has no access to the task oracle during inference, without the need to store any exemplars.

\minisection{Exemplar-free Class Incremental Learning.}
In this paper, we consider the challenging exemplar-free CIL setting.  
PASS~\cite{zhu2021prototype} aligns prototypes of new classes with those of previous tasks via augmentation with noise. SSRE~\cite{Zhu2023self} focuses on continually expanding the model's representations to accommodate new classes. FeTrIL~\cite{petit2023fetril} translates old prototype features using the difference between old and new prototypes to create pseudo-features of old data in the new space. In FeCAM~\cite{goswami2024fecam}, the heterogeneous distribution of features in class-incremental learning is exploited using anisotropic Mahalanobis distance instead of Euclidean metric. 

\minisection{Continual Representation Learning.}

CURL~\cite{rao2019curl} performs continual unsupervised representation learning and is based on a variational autoencoder whose latent space can be sampled to replay samples of old classes. 
CURL is applicable only to small-sized image datasets, requires task boundaries, and is evaluated only on simple classification task. Using self-supervised learning to enhance the learning process of a sequence of supervised tasks was introduced in prior works~\cite{zbontar2021barlow,zhu2021prototype}. Their aim is not to learn directly from unlabeled data, but rather to leverage self-supervised learning to enrich the feature representation further. Similarly, several studies~\cite{gallardo2021self,cha2021co2l,madaan2022lump,caccia2022special,ostapenko2022foundational} have used pre-trained models with self-supervision to enhance incremental average classification accuracy through techniques like data augmentation, distillation, and even exemplars. Furthermore, models are trained in self-supervised way in the class-incremental learning (CL) setting without relying on exemplars, as seen in~\cite{fini2022self,gomez2022continually}. 

Continual unsupervised learning can be straightforwardly extended to address the challenge of semi-supervised continual learning, where training phases abstain from utilizing labels entirely, similar to the approach in CaSSLe~\cite{fini2022self}. However, a limitation of this approach is mitigated by NNCSL~\cite{kang2022soft}, which builds upon the PAWS~\cite{assran2021semi} for continual learning. Nonetheless, it is worth noting that these methods typically rely on exemplars to achieve satisfactory performance.

\minisection{Drift estimation.}
An obstacle to exemplar-free methods is the drift of backbone features, which can result in catastrophic forgetting. 
Semantic drift compensation (SDC)~\cite{yu2020semantic} addresses \emph{drift compensation} by computing the drift of features using new task data and approximating the drift of old class prototypes based on this information. 
Subsequent work~\cite{toldo2022bring} extended SDC by also considering feature drift. Elastic feature consolidation (EFC)~\cite{magistri2024elastic} regularizes the drift in a direction highly relevant to the previous task using the empirical feature matrix. Recently, Adversarial drift compensation~\cite{goswami2024resurrecting} was proposed to generate adversarial samples from new task data which behaves as pseudo-exemplars and are used for estimating drift of old classes. 
Our method shares a similar objective of estimating feature backbone drift, but we propose a different approach by suggesting the use of a dedicated network for drift compensation.

Memory efficient CIL through feature adaptation (MEA)~\cite{iscen2020memory} addresses the problem of feature drift. It stores the features after the backbone and corresponding labels. 
A learned mapping is applied to translate stored features into the new space to compensate for drift. Unlike our method, MEA requires labels of features, making it unsuitable for unsupervised and semi-supervised settings. Additionally, MEA stores an extensive amount of features for all old classes, leading to higher memory requirements.

\section{Preliminaries}

\subsection{The Continual Learning Problem}
We consider a class-incremental learning setup where a model learns the dataset $D$ split into a sequence of tasks $T=\{t_1, ...,t_s\}$. During the training of each task, the model only has access to the data in the split $t$ (exemplar-free setup). Each task contains a set of $m$ classes $C^t=\{c^t_1,c^t_2,...,c^t_m\}$  and there is no overlap between classes of different tasks: $C^t\cap C^s=\emptyset$ for $t\neq s$.

Samples inside a task $t$ can contain different data depending on the learning method. For supervised learning, each sample is a pair $(x_i,y_i)$ where $x_i$ is an image of class $y_i\in C^t$. In the semi-supervised setting, $L$ samples have an assigned label $y_i$ and $U$ samples are unlabeled images $x_i$ (usually $L_t<<U_t$).

\subsection{Motivation}
\subsubsection{Not all forgetting is catastrophic.}
Semantic feature drift, as described in prior work~\cite{yu2020semantic}, manifests when a model learns sequentially. In continual learning, the prototypes computed for task $t$ may lose relevance for future tasks because the position of class $c$ after training on task $t$ differs from that in task $t-1$. This phenomenon is independent of any learning paradigm (whether supervised or unsupervised) and impacts all non-frozen models during incremental training.

\begin{figure}[tb]
     \centering        
     \includegraphics[width=0.9\textwidth]{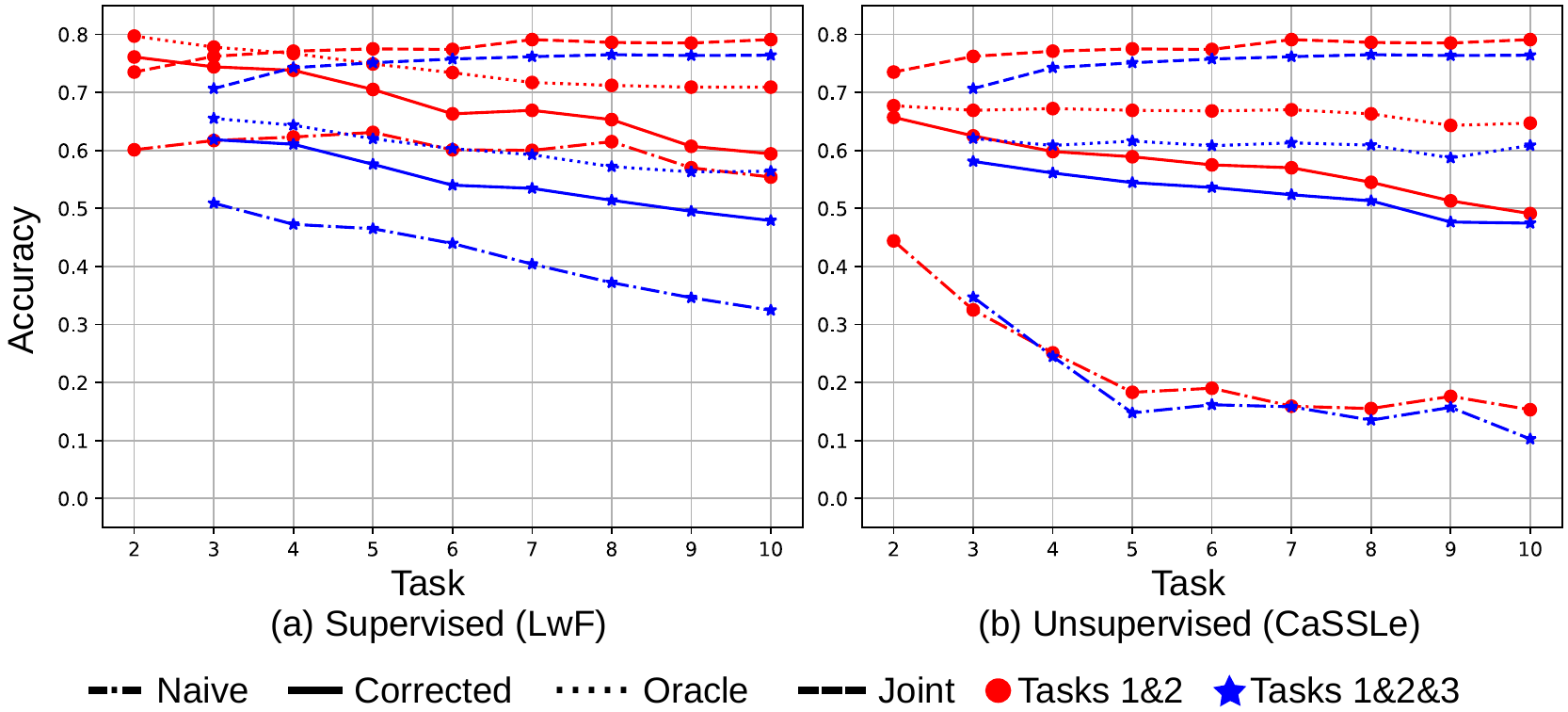}
     \caption{Last task accuracy of class-prototype accumulation strategies using an NCM classifier in the 10-task CIFAR-100 scenario. The figure row shows regularization-based methods (a) LwF and (b) CaSSLe. Here we show result for two settings: Tasks 1\&2 shows how the performance on the first two tasks evolves while incrementally training all tasks. Analogously, Tasks 1\&2\&3 shows how the performance on the first three tasks evolves over training of all ten tasks.}
       
     \label{fig:drift}
\end{figure}

To demonstrate the impact of feature drift on Class-Prototype (CP) methods, we conduct experiments using a ResNet-18 backbone trained in a Cold Start setting on 10-task CIFAR-100 and evaluate the performance of classes from the initial 3 tasks after training on all tasks. 
Fig.~\ref{fig:drift}(a) illustrates the backbone trained with the supervised continual Learning without Forgetting (LwF) strategy~\cite{li2017learning}, while Fig.~\ref{fig:drift}(b) shows the backbone trained with the unsupervised continual learning (CaSSLe) method~\cite{fini2022self}. We use a Nearest Class Mean (NCM)~\cite{rebuffi2017icarl} classifier to predict the labels of the test set for the classes encountered up to each task $t$. 

In Fig.~\ref{fig:drift}, we show the NCM accuracy using four types of prototypes:
 \begin{itemize}
     \item \textbf{Naive prototypes} ($p_{naive}$): These are prototypes without any update or correction. More specifically, prototypes of classes $C^t$ are computed at the end of $t$ and used as is afterward.
     \item \textbf{Corrected prototypes} ($p_{corrected}$): At the end of each task, all stored prototype positions are corrected using LDC (see Section~\ref{sec:LDC}), without the use of exemplars. This correction strategy is the main focus of this paper.
     \item \textbf{Oracle prototypes} ($p_{oracle}$): The true position of all old class prototypes 
     is computed by utilizing all the training data from old tasks at the end of $t$.
     \item \textbf{Jointly trained prototypes} ($p_{joint}$): The prototypes are computed using a model which is based on incremental joint training. This serves as the upper bound.
     
 \end{itemize}

We observe that while a substantial gap in performance exists between the $p_{naive}$ and $p_{joint}$ accuracy in the two plots, which is generally referred to as catastrophic forgetting, a major part of this forgetting can be mitigated using $p_{oracle}$. This indicates that a significant amount of forgetting can be attributed to the drift in class prototypes which can be reduced by correcting the position of old prototypes. For example, for the case of task 1 and 2 (red lines) we can observe that the features that are discriminative for the task 1\&2 are present in the backbone and have not been significantly impaired by the learning of subsequent tasks.
However, regularization methods (like LwF and CaSSLe) are not enough for mitigating forgetting. This motivates us to explore drift compensation. We will demonstrate that correcting the prototypes with our method LDC improves the performance of $p_{naive}$ to $p_{corrected}$ by big margins (around 38\% for tasks 1\&2\&3 in unsupervised setting).

\subsubsection{Limitations of Semantic Drift Compensation.}

The closest methods to LDC rely on SDC. However, this method has an important weakness: it assumes a fixed local transformation for drift (namely locally it can be captured by translations). If semantic drift follows a different type of transformation, such as scaling, this assumption will lead to errors in the estimated drift.

\begin{figure}[tb]
     \centering        
     \includegraphics[width=1.0\textwidth]{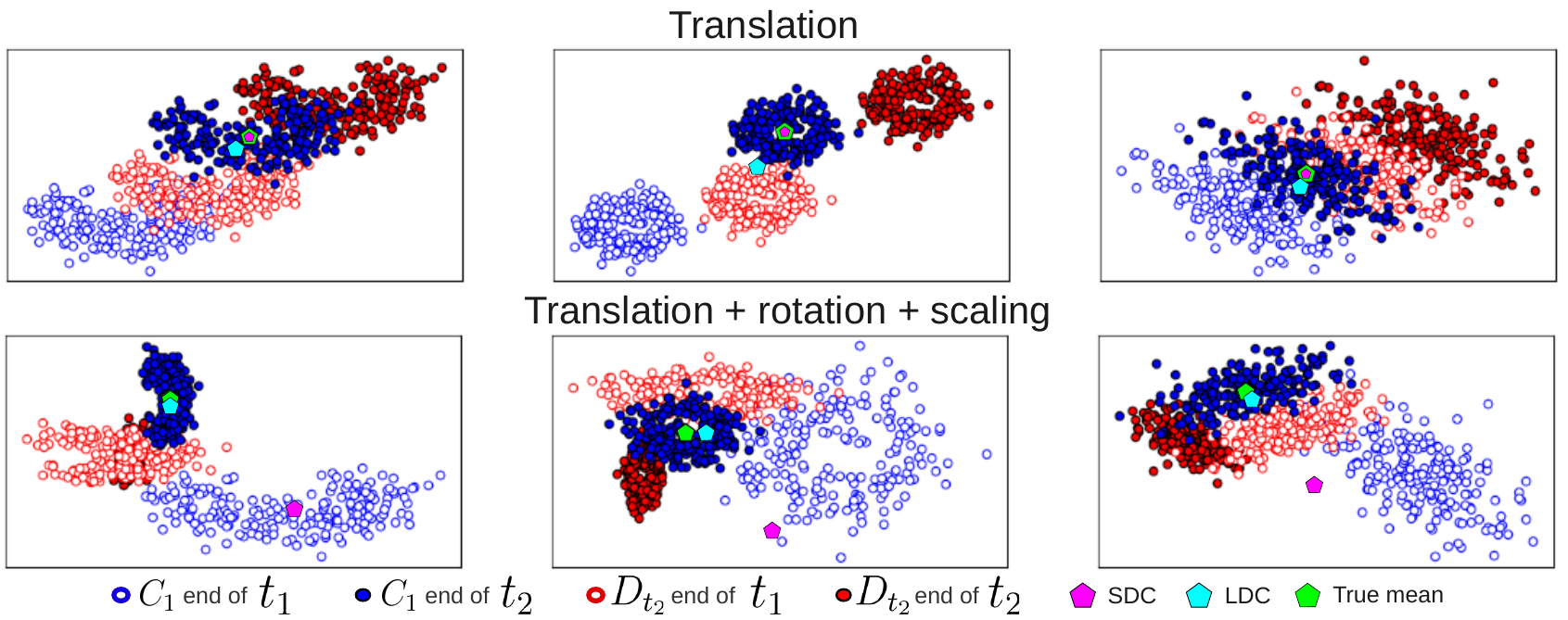}
     \caption{Feature drift estimation after applying random translations, rotations and scaling on three sample 2D distributions. We aim to estimate the true mean of $C_1$ at the end of $t_2$ using $D_{t_2}$. SDC assumes locally that transformations can be captured by translations. LDC can handle rotation and scaling in feature space. Note that LDC (\textcolor{cyan}{cyan}) more accurately approximates the real distribution mean (\textcolor{green}{green}).}
     
     \label{fig:sdc_vs_ldc}
\end{figure}

To illustrate this point, Fig.~\ref{fig:sdc_vs_ldc} presents a toy example showcasing translations, rotations, and scaling applied to three sample distributions. In this experiment, we aim to estimate the true mean of $C_1$ after it has drifted by using the transformation observed in $D_{t_2}$ as a reference. When only translations are applied (first row), SDC perfectly estimates the distribution's drift. However, under transformations that include rotations and scaling (second row), SDC incorrectly approximates the true mean of the distributions.

\subsection{Learnable Drift Compensation}
\label{sec:LDC}
\begin{figure}[t]
 \centering  
 \includegraphics[width=\textwidth]{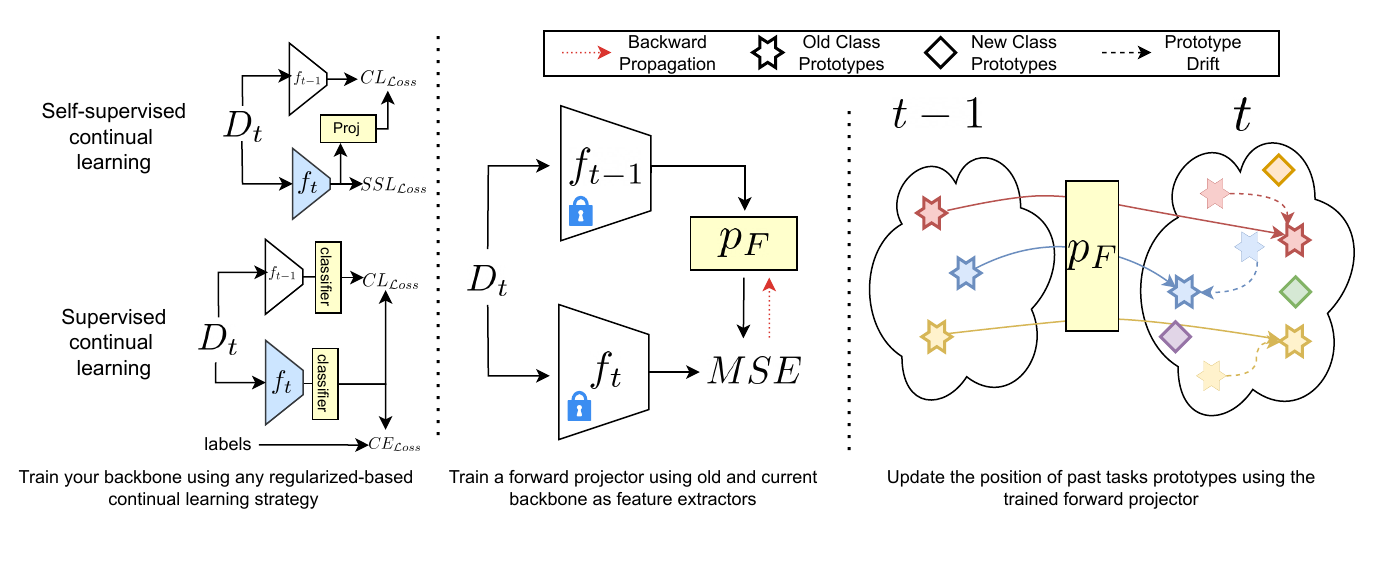}
 \caption{\textbf{Learnable Drift Compensation}. We train the model on the current task data using regularization-based, self-supervised or supervised continual representation learning methods. After training the new feature extractor $f_t$, we learn a forward projector $p_F$ by minimizing the mean squared error between the projected features from $f_{t-1}$ and the features from $f_t$. We use the learned projector to compensate for the drift of the old class prototypes in the new feature space.}
 \label{fig:LDC}
\end{figure}

LDC is applied at the end of each training session $t$. Let $f_\theta$ be a feature extractor parameterized by $\theta$. To correct the semantic drift, we learn a forward projector $p_F^{t}$ to map the features from the old feature space of $f_\theta^{t-1}$ to the new feature space of $f_\theta^{t}$. In order to learn the drift in the feature space, we use the data of the current task $D_t$ and minimize the mean squared error (MSE) between the projected features of $f_\theta^{t-1}$ and $f_\theta^{t}$. Hence, the loss function to learn $p_F^{t}$ is:
\begin{equation}
\label{eq:LDC_loss}
    \mathcal{L} = \frac{1}{N}\sum_{i=1}^N(p_F^{t}(f_\theta^{t-1}(x_i))-f_\theta^{t}(x_i))^2,
\end{equation}
where $f_\theta^{t}$ and $f_\theta^{t-1}$ are frozen.
After this optimization, the old prototypes $P_{t-1}^c$ for all classes seen till task $t-1$ are updated using Eq.~(\ref{eq:LDC}): 
\begin{equation}
\label{eq:LDC}
    P_t^c=p_F^{t}(P_{t-1}^c).
\end{equation}
This process is repeated each time $f_\theta$ is trained in a new task as shown in Fig.~\ref{fig:LDC}. Hence, if we perform class-incremental learning in $T$ tasks, $T-1$ projectors are trained. We do not need to store any old images or features, or use any labels of the current data; we simply store only the old class prototypes and update them using the learned current task projector. The projector $p_F^{t}$ can be any trainable mapping function. However, we found that a linear layer performs best in this task (see section~\ref{sec:ablation}). 
We summarized the LDC algorithm in the supplementary material. 

\minisection{Nearest Class Mean (NCM) Classifier.} NCM classifier~\cite{rebuffi2017icarl,janson2022simple,goswami2024fecam,yu2020semantic} has been found to be very effective for classification in CL settings. We use a NCM classifier on top of the trained feature extractor $f_\theta^{t}$ which classifies the test images $x$ based on their distance to all the class prototypes $P_t^c$ as follows:
\begin{equation}
\label{eq:maha_ncm}
y^\ast = \argmin\limits_{y=1,\dots,Y^t}  \|f_\theta^t(x) - P_t^y\|,
\end{equation}
where $Y^t$ is the set of classes seen up to task $t$ and $P_t^y = p_F^{t}(P_{t-1}^y) \, \forall y \in Y^{t-1}$.

\subsection{Continual Training Strategies}
Our proposed method for drift compensation is general and can be combined with many existing regularization methods commonly used in CL. LDC can be applied at the end of training session $t$ with the only requirement being access to the old feature extractor $f_{t-1}$ and the old class prototypes $P_{t-1}^c$. A projector $p_F^t$ must be optimized using  Eq.~(\ref{eq:LDC_loss}) with $D_t$, and then old prototypes are updated using Eq.~(\ref{eq:LDC}) and stored along with current ones. 
Finally, a new training session begins on $D_{t+1}$ using any CL strategy (e.g., LwF). It is important to note that our method is independent of the CL strategy used to retain knowledge in the backbone, as LDC is a prototype correction method, not a CL approach. Here, we explain how to extend several existing methods with LDC. 

\minisection{Supervised CL.}
Regularization-based methods retain knowledge while learning new concepts by preventing weights or activations from drifting far from the old model. For instance, LwF~\cite{li2017learning} aims to match the output of the previous model softmax layer $h_\theta^{t-1}(x)$ with the current model outputs $h_\theta^t(x)$ using knowledge distillation~\cite{hinton2014distilling} 
as follows:
\begin{align}
    \mathcal{L} = \mathcal{L}_{ce}(h_t(x), y) + \lambda \mathcal{L}_{ce}(h_{t-1}(x), h_t(x)),
\end{align}
where $\mathcal{L}_{ce}$ refers to the cross-entropy loss and $\lambda$ denotes the regularization strength. After the training session, the prototypes $p_t^c$ of $D_t$ are computed using the labels and added to the prototype pool for NCM classification. 
This work conducts experiments using LwF as it is a well-established and flexible method in the CL community. However, any exemplar-free supervised CL method can replace the LwF method. Regularization-based methods are particularly well-suited for LDC as they already maintain a copy of the old model at each task.

\minisection{Self-supervised CL.}
Exemplar-free unsupervised CL methods are based on a projected regularization strategy at the feature level. These approaches perform self-supervised learning (SSL) in $D_t$ while preserving similarity between old-fixed $f_{t-1}(D_t)$ and new-projected features $p(f_{t}(D_t))$. PFR~\cite{gomez2022continually} project features after the backbone and CaSSLe~\cite{fini2022self} after a post-backbone layer. POCON~\cite{gomezvilla2024} uses two networks, one expert network performs SSL in $D_t$ and the other network (main) distills the knowledge from the expert and the old main network using projected features from the current main network.

The CP accumulation strategy cannot be directly applied in exemplar-free self-supervised CL due to the absence of labels during training. However, if labels become available after training, we can compute prototypes, store them, and even correct them as in the supervised case. After the SSL CL session, we train a projector using the frozen SSL feature extractors and correct the old prototype positions accordingly.

\minisection{Semi-supervised CL.}
If some labels are accessible during training, we operate within the semi-supervised learning setup. Here, we propose using LDC with exemplar-free self-supervised CL methods, which to the best of our knowledge firstly achieve the exemplar-free continual semi-supervised learning. 
To implement this, we follow a similar procedure as in the self-supervised CL case but utilize the available labels to compute prototypes. It is important to note that the performance of LDC remains independent of the number of available labels, meaning that the primary source of error would stem from the computation of the prototype, which typically exhibits low variability.
\section{Experimental Results}\label{sec:Results}
\subsection{Datasets}
We use several publicly available datasets for our experiments. CIFAR-100~\cite{krizhevsky2009learning} contains 100 classes of $32\times32$ images with 500 training and 100 testing images per class. Tiny-ImageNet~\cite{le2015tiny} contains 200 classes of $64\times64$ images with a total of 100K training and 10K testing images. ImageNet100 is a subset of one hundred classes from ImageNet~\cite{russakovsky2015imagenet} containing $224\times224$ images with a total of 130K training and 5K testing images. Stanford Cars~\cite{krause20133d} consists of 196 finegrained categories of cars and has 8144 images for training and also for testing. We equally split all the datasets into 10 tasks, which is different from the conventional Warm Start settings~\cite{zhu2021prototype,zhu2022self,goswami2024fecam,petit2023fetril,yu2020semantic}.

\subsection{Supervised Continual Learning}
\label{sec:sup}
\minisection{Baseline methods.}
Since the existing methods were originally proposed for Warm Start settings, we re-implement all methods in Table~\ref{tab:sup_results} using PyCIL~\cite{zhou2023pycil}. We implement a standard distillation baseline method, LwF~\cite{li2017learning} and also use NCM~\cite{rebuffi2017icarl} classification naively. We use SDC~\cite{yu2020semantic} with the models trained with LwF.
Note that SDC was originally proposed in Warm Start settings with feature distillation. Here, we adapt SDC for Cold Start settings with logits distillation (LwF). We include recent methods like PASS~\cite{zhu2021prototype}, FeTrIL~\cite{petit2023fetril}, FeCAM~\cite{goswami2024fecam} and EFC~\cite{magistri2024elastic} in our comparison. While LwF, PASS and EFC are continual representation learning methods, FeCAM and FeTrIL freeze the model after the first task and continually learn the classifier. We do not compare to the methods~\cite{Malepathirana2023napa,Zhu2023self,shi2023prototype,zhu2022self} proposed in Warm Start settings since they showed poor performance compared to FeTrIL and FeCAM. Following recent practice~\cite{zhou2024continual,mcdonnell2024ranpac,zhang2023slca,goswami2024calibrating} of using pre-trained ViTs for CIL, we use a ViT-B/16~\cite{dosovitskiy2021an} model pretrained on ImageNet21k~\cite{ridnik1imagenet} and present results with NCM, SDC and LDC in Table~\ref{tab:vit_cars}.


\setlength{\tabcolsep}{2pt}
\begin{table}[t]
\begin{center}

\caption{Results in supervised settings for 10 tasks on CIFAR-100, Tiny-Imagenet, and ImageNet100. The mean and standard deviation using 5 different seeds are reported. Best results are in \textbf{bold} and second-best \underline{underlined}. $\dagger$: results excerpted from~\cite{magistri2024elastic}.}
\begin{tabular}{lcc|cc|cc}
\hline
\multirow{2}{*}{\textbf{Method}} & \multicolumn{2}{c}{\textbf{CIFAR-100}} & \multicolumn{2}{c}{\textbf{Tiny-ImageNet}} & \multicolumn{2}{c}{\textbf{ImageNet100}} \\
                        & $A_{last}$      & $A_{inc}$  & $A_{last}$      & $A_{inc}$      & $A_{last}$      & $A_{inc}$     \\
\hline\hline
LwF+NCM\cite{rebuffi2017icarl} & $40.5\pm2.7$  & $56.2\pm3.6$        & $28.6\pm1.1$             & $44.2\pm1.3$            & $44.0\pm1.9$             & $64.4\pm0.4$           \\
LwF+SDC\cite{yu2020semantic} & $40.6\pm1.8$ & $56.2\pm3.0$        & $29.5\pm0.8$             & $43.8\pm1.3$            & $42.6\pm1.9$             & $\underline{65.3}\pm0.6$           \\
PASS~\cite{zhu2021prototype} & $37.8\pm0.2$ & $52.3\pm0.1$ & $31.2\pm0.4$ & $45.3\pm0.7$ & $38.7\pm0.5$ & $55.5\pm0.5$  \\
FeTrIL~\cite{petit2023fetril} & $37.0\pm0.6$             & $52.1\pm0.5$        & $24.4\pm0.6 $             & $38.5\pm1.1$            & $40.7\pm0.5$             & $56.3\pm0.8$           \\
FeCAM~\cite{goswami2024fecam} & $33.1\pm0.9$             & $48.1\pm 1.3$        & $24.9\pm0.5$             & $38.6\pm1.3$            & $42.4\pm0.9$             & $57.9\pm1.5$           \\
EFC~\cite{magistri2024elastic} $^\dagger$ & $\underline{43.6}\pm 0.7$ & $\underline{58.6}\pm 0.9$  & $\underline{34.1}\pm 0.8$  & $\textbf{48.0}\pm 0.6$ & $\underline{47.3}\pm 1.4$             & $59.9\pm 1.4$           \\
\hline
LwF+LDC & $\textbf{45.4}\pm2.8$             & $\textbf{59.5}\pm3.9$        & $\textbf{34.2}\pm0.7$             & $\underline{46.8}\pm1.1$  & $\textbf{51.4}\pm1.2$             & $\textbf{69.4}\pm0.6$          \\
\hline

\end{tabular}
\label{tab:sup_results}
\end{center}
\end{table}

\begin{wraptable}{r}{5.5cm}
\caption{Performance on Stanford Cars (10 tasks of 20 classes each) with pre-trained ViT-B/16.}
\centering
\begin{tabular}{lccccc}
\hline
Method & $t1$ & $t3$  & $t5$  & $t7$  & $t10$ \\
\hline
\hline
FT$+$NCM & $85.8$ &  $73.3$ & $ 63.2$ & $58.0$ & $53.4$ \\
FT+SDC & $85.8$ & $73.5$ & $63.5$ & $57.8$ & $53.3$ \\
FT+LDC & $85.8$ &  $76.0$ & $71.3$ & $67.0$ & $\textbf{62.9}$ \\
\hline
\end{tabular}

\label{tab:vit_cars}
\end{wraptable}

\minisection{Training procedure.}
We train a ResNet-18~\cite{he2016deep} using an SGD optimizer. For all datasets and methods we train the first task for 200 epochs and the incremental tasks for 100 epochs. In the first task for CIFAR-100, we use a learning rate of 0.1, momentum of 0.9, weight decay of 0.0005, and the learning rate is reduced by 10 after 60, 120, and 160 epochs following the default settings of PyCIL. In the incremental tasks, we use a learning rate of 0.05, which is reduced by 10 after 45 and 90 epochs. For LwF models, we use $\lambda = 10$. For LDC, we learn a linear layer using Adam optimizer and a learning rate of 0.001 for 20 epochs using all the data in the current task. We report all details about experimental settings in the supplementary materials. 
For table~\ref{tab:vit_cars}, we use the same settings and hyperparameters as~\cite{zhang2023slca} for finetuning in new tasks and learn a two-layer MLP using Adam optimizer and a learning rate of 0.005 for 100 epochs.

\minisection{Results.} We report the last task accuracy as well as the average incremental accuracy with $5$ random seeds and class sequences in Table~\ref{tab:sup_results}. We observe that while the recently proposed methods like PASS~\cite{zhu2021prototype}, FeTrIL~\cite{petit2023fetril}, and FeCAM~\cite{goswami2024fecam} performed well in Warm Start settings, LwF with an NCM classifier outperformed these methods in Cold Start settings. While SDC was effective in the originally proposed Warm Start settings, we see that with higher feature drift in our settings, SDC does not improve the performance in most cases. We show that LDC with LwF outperforms all these baselines significantly across all datasets. In CIFAR-100, LDC outperforms SDC by $4.8\%$ and by $3.3\%$ on last task and incremental accuracy respectively. 
For Tiny-ImageNet, LDC performs similar to the recent EFC method which is also designed for the cold start setting. However, on the more challenging ImageNet100 LDC improves over EFC by a large margin of almost 4\% on last task accuracy.
Finally, using pre-trained ViTs on Stanford Cars, we observe that LDC improves over SDC by 9.6\% after the last task, see Table~\ref{tab:vit_cars}.

\setlength{\tabcolsep}{6pt}

\begin{table}[t]
\begin{center}
\caption{Results in semi-supervised settings for $10$ tasks on CIFAR-100. The mean and standard deviation using $3$ different seeds are reported. The best results for the proposed setting are in \textbf{bold}.}
\label{tab:semicifar}
\begin{tabular}{lccccc}
\hline
\textbf{Method}     & \textbf{Ex Free} & \textbf{Semi-Sup} & & \textbf{100\%} \\
\hline\hline
LwF+NCM        & \cmark & \xmark & & 40.5$\pm$2.7 \\
PASS       & \cmark & \xmark & & 37.8$\pm$0.2 \\
\hline\hline

\rule{0pt}{3ex} 
& & & \textbf{0.8\%} & \textbf{5\%} & \textbf{25\%} \\
\hline
PASS       & \cmark & \cmark & 5.91 & 15.13 & 15.84 \\
\hline
NNCSL (500)      & \xmark & \cmark & 27.4$\pm$0.51 & 31.4$\pm$0.40 & 35.3$\pm$0.30\\
NNCSL (0)       & \cmark & \cmark  & $\phantom{2}$8.7$\pm$0.63 & $\phantom{2}$9.0$\pm$0.42 & 9.32$\pm$0.39 \\
\hline
PFR+naive    & \cmark & \cmark &  20.5$\pm$0.44 &  24.8$\pm$0.35 & 25.7$\pm$0.12 \\
CaSSLe+naive & \cmark & \cmark & 19.4$\pm$0.57 & 23.2$\pm$0.21 & 23.6$\pm$0.13 \\
POCON+naive  & \cmark   & \cmark &  20.1$\pm$0.23 & 21.7$\pm$0.27 & 22.4$\pm$0.31\\
\hline
PFR+SDC    & \cmark & \cmark & 16.2$\pm$0.66 & 22.5$\pm$0.57 & 24.8$\pm$0.12 \\
CaSSLe+SDC & \cmark & \cmark & 22.1$\pm$0.69 & 25.8$\pm$0.09 & 26.5$\pm$0.04 \\
POCON+SDC  & \cmark   & \cmark   & 21.8$\pm$0.82 & 25.7$\pm$0.23 & 26.7$\pm$0.09    \\
\hline
PFR+LDC    & \cmark & \cmark & 18.6$\pm$0.20 & 29.3$\pm$0.12 & 33.4$\pm$0.06\\
CaSSLe+LDC & \cmark & \cmark & \textbf{27.0}$\pm$0.60  & \textbf{32.8}$\pm$0.30 & \textbf{35.0}$\pm$0.21  \\
POCON+LDC  & \cmark     & \cmark  & 26.1$\pm$0.74 & 31.1$\pm$0.62 & 33.8$\pm$0.25    \\
\hline
\end{tabular}
\end{center}
\end{table}

\subsection{Semi-supervised Continual Learning}\label{sec:semi}
\minisection{Baseline methods.}
Since the proposed method is the first exemplar-free semi-supervised CL method, we can only compare it against drift correction methods (SDC). Nevertheless, we present results for the most close approaches. The naive approach achieves the lowest possible accuracy, no prototype correction is performed. SDC~\cite{yu2020semantic} is used as a comparison for drift correction. PASS~\cite{zhu2021prototype} is not a semi-supervised method, but we present our results with all labels (as a reference) and with reduced training data corresponding to the proposed label sets. NNCSL~\cite{kang2022soft} is a semi-supervised method that uses exemplars; we present here results with the same number of exemplars proposed by the authors (as a reference) and a reduced version.
Removing exemplars from exemplar-based methods or reducing data to comply with our setup is not fair, and we present these results as a reference to showcase the performance of our approach. We use the code and hyperparameters provided by the authors.

\minisection{Training procedure.}
We train a ResNet-18~\cite{he2016deep} using PFR~\cite{gomez2022continually}, CaSSLe~\cite{fini2022self}, and POCON~\cite{gomezvilla2024} as SSL CL strategies. We use the same code, training procedure, and hyperparameters provided by the authors in each method.
At the end of each task, we compute the prototypes using the available labels: $0.8\%$, $5\%$, $25\%$ for CIFAR-100 and $1\%$, $5\%$, $25\%$ for ImageNet100. For LDC, we train a linear projector using the Adam optimizer (learning rate $5e-3$) for 100 epochs using all the data in the task. We report the mean last task accuracy in the test set using an NCM classifier.

\minisection{Results.}
Table~\ref{tab:semicifar} presents the results of 10 task split of CIFAR-100. LDC presents the best results for all the pure semi-supervised exemplar-free settings. For $5\%$ and $25\%$ settings, it is reaching the same accuracy as the exemplar-based method NNCSL. Furthermore, LDC is only $3\%$ below the performance of the fully supervised version of PASS. LDC surpasses SDC by at least $7\%$ in all the base SSL CL methods in the pure semi-supervised exemplar-free setting.

Table~\ref{tab:semiin} presents the results of 10 task split of ImageNet100. As in CIFAR-100, LDC performs best for the proposed setting, surpassing SDC by at least $7\%$ in all the SSL CL backbones. However, in this dataset, the replay-based method NNCSL is better with the same number of exemplars proposed by the authors. Nevertheless, a reduction in the number of exemplars in NNCSL leads to a significant performance drop, ending below our proposed exemplar-free method.

\begin{table}[t]
\begin{center}
\caption{Results in semi-supervised settings for $10$ tasks on ImageNet100. The mean and standard deviation using $3$ different seeds are reported. The best results for the proposed setting are in \textbf{bold}.}
\label{tab:semiin}
\begin{tabular}{lccccc}
\hline
\textbf{Method}     & \textbf{Ex Free} & \textbf{Semi-Sup} & & \textbf{100\%} \\
\hline\hline
LwF+NCM        & \cmark& \xmark & & 44.0$\pm$1.9 \\
PASS       & \cmark & \xmark & & 38.7$\pm$0.5 \\
\hline

\rule{0pt}{3ex} 
& & & \textbf{1\%} & \textbf{5\%} & \textbf{25\%} \\
\hline\hline
PASS       & \cmark & \cmark & 1.76 & 10.36 & 23.9 \\
\hline
NNCSL (5120)      & \xmark & \cmark  & 35.4$\pm$0.21 & 41.6$\pm$0.12 & 47.1$\pm$0.08\\
NNCSL (500)       & \xmark & \cmark  & 9.93$\pm$0.22 & 15.1$\pm$0.15 & 16.6$\pm$0.1 \\
\hline
PFR+naive    & \cmark & \cmark &   19.6$\pm$0.16 & 21.8$\pm$0.38 & 22.0$\pm$0.12 \\
CaSSLe+naive & \cmark & \cmark &  25.37$\pm$0.27 & 26.9$\pm$0.25 & 27.36$\pm$0.11 \\
POCON+naive  & \cmark   & \cmark &  24.2$\pm$0.32 & 26.1$\pm$0.22 & 25.8$\pm$0.20\\
\hline
PFR+SDC    & \cmark & \cmark &  17.2$\pm$0.55 & 19.1$\pm$0.16 & 19.8$\pm$0.08\\
CaSSLe+SDC & \cmark & \cmark & 22.2$\pm$0.03 & 25.9$\pm$0.23 & 26.6$\pm$0.01  \\
POCON+SDC  & \cmark   & \cmark &  21.8$\pm$0.08 & 25.1$\pm$ 0.30 & 25.7$\pm$0.03\\
\hline
PFR+LDC    & \cmark & \cmark &  26.5$\pm$0.09 & 32.4$\pm$0.35 & 33.5$\pm$0.07 \\
CaSSLe+LDC & \cmark & \cmark &  \textbf{29.5}$\pm$0.58 & \textbf{33.9}$\pm$0.15 & \textbf{35.8}$\pm$0.39 \\
POCON+LDC  & \cmark     & \cmark & 28.3$\pm$0.37 &  33.2$\pm$0.30 & 34.0$\pm$0.28 \\
\hline
\end{tabular}
\end{center}
\end{table}

\subsection{Ablation Studies}\label{sec:ablation}
In this section we report on a range of ablation studies for LDC. Unless specified, the training and evaluation procedure are the same as in Sections~\ref{sec:sup}~and~\ref{sec:semi}. We provide more ablation experiments on comparison of LDC with NME and using stored features for learning the projector, in the supplementary materials.

\minisection{Architecture of projector.}
To ablate the complexity of the proposed projector $p_F$,   
Table~\ref{tab:projector} presents experiments for LwF+LDC(supervised CL) and CaSSLe+LDC (semi-supervised CL - $25\%$ of labels). We show LDC performance with a linear layer, with and without bias, with an activation layer ReLU after the linear layer and also using a two-layer MLP. We observe that a single linear layer works better in both settings.

\begin{table}[t]
\centering
\caption{Ablation study for different complexities of the proposed projector $p_F$.}
\begin{tabular}{lcccc}
\label{tab:projector}
\\
\hline
\textbf{Method} & \textbf{Linear} & \textbf{Linear + bias} & \textbf{Linear + ReLu} & \textbf{MLP} \\
\hline\hline
LwF + LDC & 45.4$\pm$2.8  & 45.2$\pm$3.0 & 41.2$\pm$2.1 & 43.7$\pm$2.3 \\
CaSSLe + LDC & 35.0$\pm$0.2  & 34.3$\pm$0.2 & 29.1$\pm$0.5  & 34.9$\pm$0.4  \\
\hline
\end{tabular}
\end{table}

\begin{figure}[t]
 \centering  
 \includegraphics[width=\textwidth]{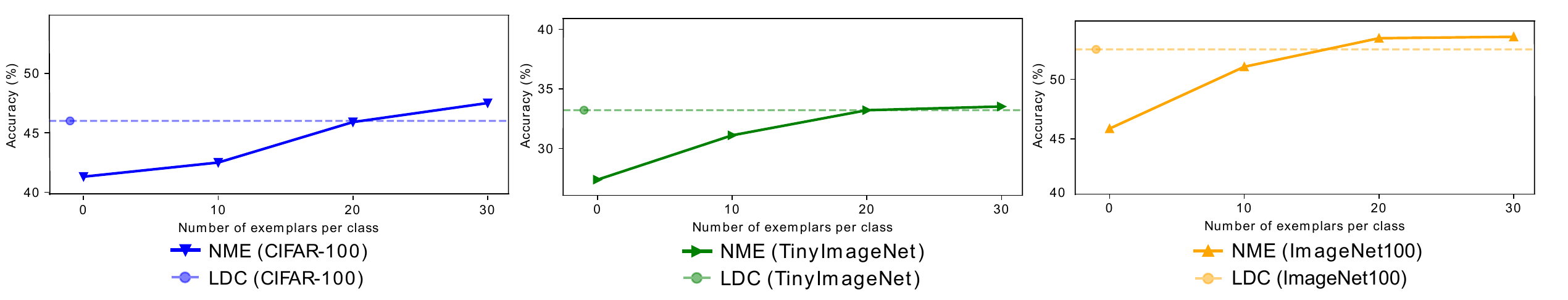}
 \caption{Comparison of LDC with NME for varying memory size in the supervised settings on CIFAR-100, Tiny-ImageNet and ImageNet100.}
 \vspace{-10pt}
\label{fig:nme}
\end{figure}

\minisection{Exemplars versus LDC for prototype correction.}
We compare the performance of LDC against Nearest-Mean of Exemplars (NME) which directly uses exemplars for old class prototype compensation and is explored in exemplar-based methods~\cite{douillard2020podnet,rebuffi2017icarl}. While it is easy to predict the updated positions of the old prototypes using original samples of old classes by NME, it is very challenging without exemplars in our settings. We compare the last task accuracy of LDC (same setting as in Table~\ref{tab:sup_results}, with LwF+NCM) with NME which stores a memory of old class samples. We show in Fig.~\ref{fig:nme} that LDC is on par with using 20 exemplars per class for NME for CIFAR-100 and Tiny-ImageNet. For ImageNet100, LDC is marginally less than NME with 20 exemplars per class. This demonstrates that LDC is able to effectively correct the prototypes without access to exemplars and yet is on par with NME using exemplars.

\begin{figure}[t]
 \centering  
 \includegraphics[width=\textwidth]{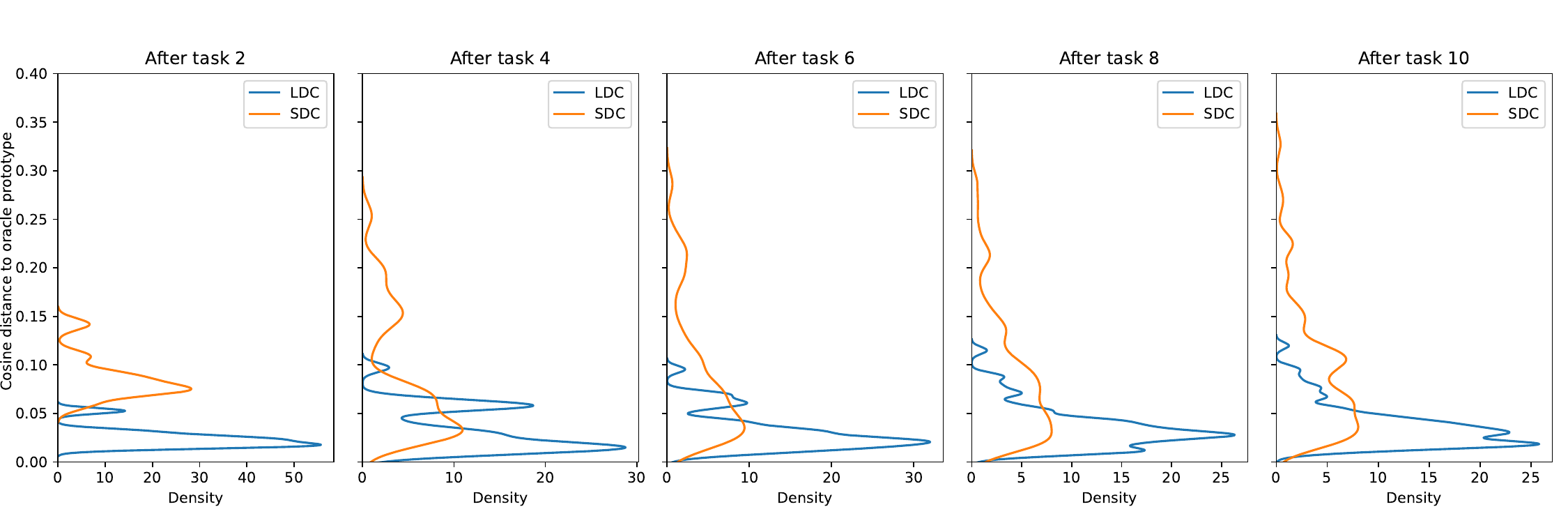}
 \caption{Drift compensation analysis using SDC and LDC in supervised settings on CIFAR-100. We compute the cosine distance between the updated and oracle prototypes. We plot the distributions of the cosine distances for all old classes using both methods after alternate tasks. While SDC fails to estimate the prototypes closer to the oracle (with increasing distance for many classes), LDC predictions are very close to the oracle prototypes (mean of distributions using LDC is close to 0 even after the last task and has low standard deviation).}

 \label{fig:kde}

\end{figure}

\minisection{Drift Compensation Analysis.}
In order to visualize the drift compensation using LDC, Fig.~\ref{fig:kde} presents a comparison between SDC and LDC. We update all the old prototypes using both SDC and LDC and then compute the cosine distance of the updated prototype to the oracle prototype (computed using all old data for analysis) in the new feature space. We plot the distributions over all old class prototype distances after alternate tasks till the lats task. We observe that SDC struggles to estimate the prototypes even after the first increment (task 2) and is centered around distance of 0.1. As we move to the last task, SDC gets even worse with a higher deviation which implies increasing distance of updated prototypes from their oracle positions. On the other hand, LDC is more robust and the distribution is centered between 0 and 0.05 even after the last task, with much lower deviation. This analysis suggests that the LDC predicts the old prototypes very close to their oracle positions and thus better tracks the prototypes position when moving from one feature space to another.

\section{Conclusions}
We showed that performance degradation in exemplar-free CP accumulation methods is largely due to feature drift. We proposed a simple yet effective prototype correction method called Learnable Drift Compensation (LDC). LDC learns a projector that maps between the feature spaces of the consecutive tasks using only the data available at each task along with current and previous model. The trained projector is used at the end of each task to correct the stored old prototype positions. We demonstrate the efficacy of LDC in correcting the drift of any moving backbone by using it on top of already established exemplar-free CL strategies. Extensive experiments on CIFAR-100, Tiny-ImageNet, and ImageNet100 show that LDC outperforms previous drift correction methods in supervised and semi-supervised settings. 

\minisection{Limitations.} 
While the learned projection $p_f$ between tasks is effective for correcting drift, it's not flawless. The gap between $p_{corrected}$ and $p_{oracle}$ in Fig.~\ref{fig:drift} highlights this difference. Without exemplars, $p_f$ only learns the projection from current task data. Consequently, due to biases in the current data, updates to some old class prototypes may not be perfect. We would like to explore methods to further minimize or eliminate this gap in our future work.

\newpage

\minisection{Acknowledgments.} We acknowledge projects TED2021-132513B-I00 and PID2022-143257NB-I00, financed by MCIN/AEI/10.13039/501100011033 and FSE+ by the European Union NextGenerationEU/PRTR, and by ERDF A Way of Making Europa, and the Generalitat de Catalunya CERCA Program. Bartłomiej Twardowski acknowledges the grant RYC2021-032765-I.



%
%
\bibliographystyle{splncs04}
\bibliography{main}

\newpage
\appendix
\section*{Supplementary Material}
\section{Learnable Drift Compensation pseudocode}
\begin{algorithm}
	\SetKwInOut{Input}{input}\SetKwInOut{Output}{output}
	\SetKwData{minibatch}{minibatch}
	\SetKwData{trainingbatches}{training\_batches}
	\SetKwData{task}{task}
	\SetKwData{prototype}{prototype}
	\SetKwFunction{Freeze}{freeze}
	\SetKwFunction{Minimize}{minimize}
	\SetKwFunction{Mse}{MSE}
	\SetKwInOut{Phase}{Phase 1}
	\SetKwInOut{Phaseb}{Phase 2}
	\SetKwInOut{Phasec}{Phase 3}
	\caption{Learnable Drift Compensation}\label{alg:LDC}
	At the end of each training session $t$ \\
	\Input{$D_t$;$f_\theta^{t-1}$;$f_\theta^{t}$;Prototype pool of old task classes $p^C$}
		\Phase{Train feature extractor $f_\theta^{t}$ in $D_t$ using any learning strategy}
		\Phaseb{Train forward projector $p_F^t$ using $D_t$; $f_\theta^{t-1}$; $f_\theta^{t}$}
		\Begin{
			\Freeze{$f_\theta^{t-1}$,$f_\theta^{t}$}\\
			
			\For{\minibatch in \trainingbatches}{
				\Minimize{\Mse{$p_F^t(f_\theta^{t-1}(\minibatch))$,$f_\theta^{t}(\minibatch)$}}
			}
			
		}
		\Phasec{Update old prototypes with $p_F^t$}
		\Begin{
			\For{\prototype in $p^C$}{
				$p^{\prototype}=p_F^t(p^{\prototype})$
			}
		}
\end{algorithm}

\section{Training Settings}

\subsection{Supervised CL settings}
We implemented existing methods like LwF~\cite{li2017learning} (with NCM~\cite{rebuffi2017icarl} and SDC~\cite{yu2020semantic}), PASS~\cite{zhu2021prototype}, FeTrIL~\cite{petit2023fetril} and FeCAM~\cite{goswami2024fecam} and optimized them for small-start settings. We use the PyCIL~\cite{zhou2023pycil} framework. Here we share the details of the settings for reproducibility. 

For CIFAR-100, we use the CIFAR-10 policy augmentations from PyCIL and use it for all the methods to have a fair comparison. For all datasets, we follow the common practice of using random crop and random horizontal flip for the training set.

\minisection{LwF:} For all datasets in the first task, we follow PyCIL and use a learning rate of 0.1, momentum of 0.9, weight decay of 0.0005 and train with a batch size of 128 for 200 epochs. The learning rate is reduced by 10 after 60 , 120 and 180 epochs. For all new tasks in CIFAR-100 and ImageNet100, we use a learning rate of 0.05. For TinyImageNet, we use a learning rate of 0.001. We train for 100 epochs and reduce the learning rate by 10 after 45 and 90 epochs. The temperature scale is set to 2. The LwF regularization strength is set to 10 for CIFAR-100 and TinyImageNet and 5 for ImageNet100.
We use the models trained with LwF and use them with a NCM classifier and SDC. For FeTrIL and FeCAM, we train the first task in the same way as LwF.

\minisection{SDC:} For SDC~\cite{yu2020semantic}, we use $\sigma=0.3$, which is the standard deviation of the Gaussian kernel for estimating the weights of the drift vectors. For ImageNet100, we set $\sigma=1.0$.

\minisection{PASS:} Following the PyCIL implementation of PASS, we set $\lambda_{fkd} = 10$ and $\lambda_{proto} = 10$.

\minisection{FeTrIL:} For all incremental tasks, we follow the same settings as the FeTrIL implementation from PyCIL~\cite{zhou2023pycil}.

\noindent \textbf{FeCAM:} FeCAM trains the first task and use the frozen model for all subsequent tasks. FeCAM use the stored prototypes and distribution statistics (covariance matrix) for all old classes. Following the implementation from~\cite{goswami2024fecam}, we use the covariance shrinkage hyperparameters of (1,1) and 0.5 as the Tukey's normalization value.

\subsection{Self-supervised CL settings.}
We use the existing CL exemplar-free methods namely PFR~\cite{gomezvilla2022continually}, CaSSLE~\cite{fini2022cassle}, and POCON~\cite{gomezvilla2024}. We utilize the code provided for each method. Here we include additional details of the settings for reproducibility. 

For each dataset, we follow the image augmentation pipeline of SimCLR~\cite{chen2020simple}. All the methods use Barlow Twins~\cite{zbontar2021barlow} as base self-supervised learning strategy.

\minisection{PFR:} We conduct our experiments with $500$ epochs per task on CIFAR100 and $400$ for ImageNet100. As for CL hyperparameters, $\lambda=25$ for both datasets; the learning rate, optimizer, and training schedules are the same as proposed by the authors.

\minisection{CaSSLe:} We use the same training epochs as PFR. We strictly follow the hyperparameters shared by the authors.

\minisection{POCON:} For both datasets, we employ the configuration CopyOP. The hyperparameters setup is the same as the authors proposed for the 10-task partition of each dataset.

\section{Additional Ablation Experiments}

\minisection{Feature Storage.}
In this work, we update only prototypes (one sample mean per class), but one can argue that storing a set of features per class is better than a single class mean. In Table~\ref{tab:feat_stats}, instead of projecting a prototype per class, we project $N$ stored features and use them to compute the class prototype at the classification stage. The results show that storing and projecting features do not present a significant advantage over projecting the class means.

\begin{table}[t]
	\centering
	\caption{Accuracy comparison storing and projecting $N$ features per class against storing and projecting only class prototypes.}
	\begin{tabular}{cccccc}
		\label{tab:feat_stats}
		\\
		\hline
		\textbf{Method} & \textbf{Mean} & $\boldsymbol{N=5}$ & $\boldsymbol{N=50}$ & $\boldsymbol{N=100}$ & $\boldsymbol{N=500}$ \\
		\hline\hline
		CaSSLe + LDC & $35.0\pm0.2$  & $21.1\pm0.01$ & $32.9\pm0.03$ & $33.9\pm0.02 $& $35.05\pm0.04 $\\
		\hline
	\end{tabular}
\end{table}

\end{document}